\documentclass[sigconf,natbib=true,anonymous=false,review=false]{acmart}

\copyrightyear{2025}
\acmYear{2025}
\acmConference[ICTIR '25]{Proceedings of the 2025 International ACM SIGIR Conference on Innovative Concepts and Theories in Information Retrieval (ICTIR)}{July 18, 2025}{Padua, Italy}
\acmBooktitle{Proceedings of the 2025 International ACM SIGIR Conference on Innovative Concepts and Theories in Information Retrieval (ICTIR) (ICTIR '25), July 18, 2025, Padua, Italy}\acmDOI{10.1145/3731120.3744584}
\acmISBN{979-8-4007-1861-8/2025/07}

\usepackage{arydshln}

\usepackage{enumitem}
\newlist{inlinelist}{enumerate*}{1}
\setlist*[inlinelist,1]{%
  label=(\roman*),
}
\usepackage{xspace}
\usepackage{amsmath}
\usepackage{multicol}
\usepackage{multirow}
\usepackage{xcolor}
\usepackage{algorithm}
\usepackage{algorithmic}
\usepackage{mathtools}
\usepackage{adjustbox}
\usepackage{subfig}
\usepackage{graphicx}

\newcommand{\ourframework}{\textsc{IUM}\xspace}
\newcommand{\rlem}{\textsc{Offline IUM}\xspace}
\newcommand{\sessionlearning}{\textsc{Online IUM}\xspace}
\newcommand{\KL}{D_{\mathrm{KL}}}

\title{Learning to Rank for Multiple Retrieval-Augmented Models through Iterative Utility Maximization}

\author{Alireza Salemi}
\affiliation{\institution{University of Massachusetts Amherst}
\city{Amherst}
\state{MA}
\country{United States}}
\email{asalemi@cs.umass.edu}

\author{Hamed Zamani}
\affiliation{\institution{University of Massachusetts Amherst}
\city{Amherst}
\state{MA}
\country{United States}}
\email{zamani@cs.umass.edu}

\begin{document}


\begin{abstract}
This paper investigates the design of a unified search engine to serve multiple retrieval-augmented generation (RAG) agents, each with a distinct task, backbone large language model (LLM), and RAG strategy. We introduce an iterative approach where the search engine generates retrieval results for the RAG agents and gathers feedback on the quality of the retrieved documents during an offline phase. This feedback is then used to iteratively optimize the search engine using an expectation-maximization algorithm, with the goal of maximizing each agent's utility function. Additionally, we adapt this to an online setting, allowing the search engine to refine its behavior based on real-time individual agents feedback to better serve the results for each of them. Experiments on datasets from the Knowledge-Intensive Language Tasks (KILT) benchmark demonstrates that our approach significantly on average outperforms baselines across 18 RAG models. We demonstrate that our method effectively ``personalizes'' the retrieval for each RAG agent based on the collected feedback. Finally, we provide a comprehensive ablation study to explore various aspects of our method.
\end{abstract}





\maketitle

\section{Introduction}

Search engines have been designed mainly to serve people by providing relevant results for search queries. They are typically trained at scale using learning-to-rank methods with implicit feedback gathered and refined over time through user interactions \citep{10.1145/2568388.2568413, 10.1145/3018661.3018699, 4292009}. These systems often use personalization to deliver more tailored and effective search results for users \cite{5941815, 10.1145/1321440.1321515}. With the recent rise of large language models (LLMs), along with their user-friendly chat interfaces, many users now use them for a wide range of tasks \citep{caramancion2024largelanguagemodelsvs}. Extensive studies have shown that LLMs often face challenges in tasks that require external knowledge in rapidly changing or evolving environments \citep{kilt, reml}. One method to address this limitation is to enhance LLMs by retrieving information from a knowledge source, a technique known as retrieval-augmented generation (RAG) \citep{rag, asai2023selfrag, fid}. This marks a paradigm shift in the current landscape, where humans interact primarily with LLMs, while LLMs rely on search engines to retrieve external information \citep{reml, kim2024retrievalenhancedmachinelearningsynthesis, urag}. 

With this paradigm shift, it is crucial to develop methods for optimizing and personalizing search engines to better serve their new users: the LLMs that depend on them. This need is further highlighted by prior work showing significant differences between the information preferences of humans and RAG agents \citep{erag}. In addition, different LLMs may have different information needs. For example, smaller LMs with limited reasoning capabilities might prefer supporting documents that provide explicit information, while more powerful LLMs can infer details even when the information is stated vaguely. Given these variations, optimizing engines to address the unique requirements of different LLMs is essential.

Prior work on RAG mostly focus on developing a RAG system (i.e., a retrieval and a language model) for each task. The recent work by \citet{urag} represents one of the first efforts to build a centralized search engine capable of serving \textit{multiple} RAG agents, each tasked with a distinct objective. These agents utilize different underlying LLMs and employ varying RAG techniques. To train this centralized search engine, they use its initial parameters to retrieve documents for the training queries of the agents (i.e., the LLMs utilizing the search engine). The feedback from these agents on the retrieved documents is then collected to update the parameters. This approach has a few limitations. First, it depends heavily on the initial parameters to retrieve documents for training queries. If these parameters are not well-initialized, the quality of the initial retrieval lists would be low, leading to suboptimal feedback from the LLMs. Furthermore, since the initial documents are retrieved without any prior feedback to adapt the search engine to agent needs, they may not be relevant or useful to the agents. Consequently, the feedback provided might not reflect useful documents for the individual agents, reducing the overall effectiveness of the training process.

This paper addresses these limitations by introducing an iterative approach grounded in strong theoretical foundations, designed to maximize the utility functions of RAG agents. Feedback is collected over multiple iterations in an offline training procedure to progressively optimize the search engine. In each iteration, the search engine applies the parameters optimized from the previous feedback round to retrieve new documents tailored to the specific information needs of each RAG agent. Feedback on these newly retrieved documents is then collected to refine the search engine's understanding of each agent's preferences. The personalization process adapts the search engine for each RAG agent by leveraging feedback from the prior iteration, using the agents' training queries and feedback on them. During the offline phase, feedback from all agents is aggregated to train a multi-task central search engine capable of effectively serving the diverse needs of all agents.

A natural extension to the offline optimization phase is online learning in the serving phase where the retriever provides search results for the agents on test queries. Here, the same iterative algorithm is applied to adapt the retriever based on agent feedback obtained during active serving sessions. Specifically, the retriever first processes a batch of queries from an agent, collects feedback on these queries, and then updates its parameters to improve performance on subsequent queries within the same session for this specific agent. This optimization occurs concurrently with the serving process and is not confined to pre-defined training queries. Since this optimization leverages each agent's feedback individually to further optimize itself, it enhances the personalization of the search engine for that specific agent even more, tailoring the system to better meet the unique information needs of each agent.

We evaluate our approach using diverse tasks from the Knowledge-Intensive Language Tasks (KILT) benchmark \citep{kilt}. Our evaluation includes three open-domain question answering datasets: Natural Questions (NQ) \citep{kwiatkowski-etal-2019-natural}, TriviaQA \citep{joshi-etal-2017-triviaqa}, and HotPotQA \citep{yang-etal-2018-hotpotqa}; one fact verification dataset: FEVER \citep{thorne-etal-2018-fever}; and two relation extraction and slot-filling datasets: zsRE \citep{levy-etal-2017-zero} and T-REx \citep{elsahar-etal-2018-rex}. Following \citet{urag}, we test our approach with 18 different RAG agents as the users, each performing a specific task and utilizing distinct augmentation approaches and LLMs, to serve as users of the search engine. Our results demonstrate that the proposed approach for offline iterative training of the search engine significantly outperforms the state-of-the-art baseline. Additionally, combining this offline approach with our online learning yields an even greater improvement over the baselines. We also conduct an extensive ablation study on various configurations of the proposed approaches to provide further insights into their effectiveness and impact. Furthermore, we show that our approach enhances the personalization of the search engine over time, addressing a limitation noted in \citet{urag}. We observe that the correlation between the retrieval results of the agents employing different LLMs but performing the same task is very low, indicating that the results are effectively personalized. To support future research, we have open-sourced our code and models for the community. \footnote{Our code can be found at: 
\url{https://github.com/alirezasalemi7/uRAG}}

\section{Related Work}

\subsubsection*{\textbf{Knowledge-Intensive Language Tasks (KILT)}}

Contrary to standard NLP tasks like natural language understanding \citep{glue, superglue} and question answering \citep{mccann2019the}, where the input alone is enough to perform the task, knowledge-intensive NLP tasks rely heavily on external knowledge sources to extract necessary information. \citet{kilt} introduces KILT, a benchmark designed for evaluating such tasks. KILT encompasses a variety of tasks, including open-domain question answering, fact verification, slot filling, and entity linking, providing a benchmark for knowledge-intensive tasks.

\subsubsection*{\textbf{Retrieval-Augmented Generation (RAG)}}

RAG \citep{rag} represents a framework that merges information retrieval with natural language generation to improve the quality of generated outputs by integrating external knowledge in the generation process \citep{asai2023selfrag, siriwardhana-etal-2023-improving}. Unlike traditional LLMs that rely solely on pre-trained knowledge, RAG can dynamically retrieve information from external sources via a retriever, enabling them to produce content that is more accurate \citep{reml, kim2024retrievalenhancedmachinelearningsynthesis}. This flexibility allows RAG to be applied in various domains, including knowledge grounding in textual \citep{kilt, rag, fid} and multimodal \citep{dedr, murag, kat, salemi2023pretraining}, personalization \citep{salemi2024lamp, salemi2024optimization, kumar2024longlampbenchmarkpersonalizedlongform, salemi2024comparingretrievalaugmentationparameterefficientfinetuning,salemi2025reasoningenhancedselftraininglongformpersonalized, salemi2025experteffectiveexplainableevaluation}, and reducing hallucinations \citep{agrawal2023knowledge, shuster-etal-2021-retrieval-augmentation}. The retriever in RAG plays a pivotal role as it sources the necessary information for the LLM to perform its task \citep{rag}. This is typically done using either sparse retrieval methods (e.g., TF-IDF, BM25 \citep{bm25}) or dense retrieval models (e.g., DPR \citep{dpr}, Contriever \citep{contriever}, ColBERTv2 \citep{santhanam-etal-2022-colbertv2}, E5 \citep{wang2024textembeddingsweaklysupervisedcontrastive}). The retrieved information is then utilized by the large language model to complete the task. Prominent methods in this context include In-Prompt Augmentation (IPA) and Fusion-in-Decoder (FiD) \citep{fid}. In IPA, the retrieved data is appended to the prompt, allowing the language model to incorporate it during generation. FiD encodes each retrieved document separately alongside the prompt within the encoder of an encoder-decoder architecture, combining them in the decoder to generate a cohesive answer based on the available information, as explained in \citet{fid}. 

\subsubsection*{\textbf{A Search Engine for Machines}}

Research on search engines show that successful systems rely on large-scale feedback for optimization \citep{10.1145/1148170.1148177, NIPS1999_7283518d}. With LLMs as the primary users of search engines \citep{reml, kim2024retrievalenhancedmachinelearningsynthesis}, \citet{erag} showed that the LLMs' preferences about relevance of a query and document differs from humans. \citet{urag} introduced a new problem that is training a unified search engine capable of serving multiple diverse RAG agents. They introduced uRAG, a unified ranking model designed to serve multiple RAG models while learning and optimizing based on feedback provided by these diverse RAG models. Recently, several methods have been proposed for training retrieval models tailored to LLMs, including distillation from LLMs to retrievers \citep{retriever-reader-distill, distil-fid, atlas}, end-to-end training of retrievers and LLMs \citep{sachan-etal-2021-end, StochasticRAG}, and bandit algorithms \citep{rl-retriever-train}. However, most of these approaches focus on leveraging feedback from a single LLM, with the aim of aligning the retrieval with that particular LLM \citep{urag}. Here, we introduce an approach based on iterative utility maximization to train a unified retrieval model for serving multiple RAG agents, applied in both offline and online settings \citep{hoi2018onlinelearningcomprehensivesurvey, 10.5555/2188385.2188391} to optimize the search engine for the agents.

\begin{figure*}
    \centering
    \includegraphics[width=0.8\linewidth]{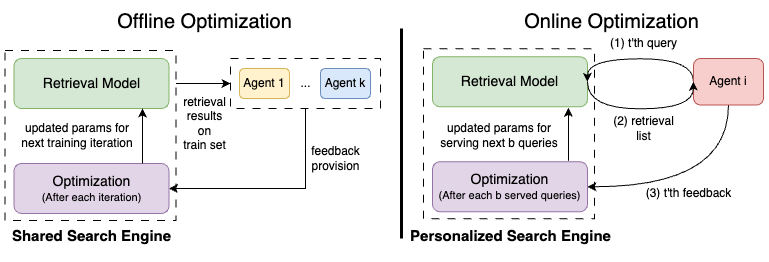}
    \vspace{-0.6cm}
    \caption{Iterative Utility Maximization (\ourframework). This framework first iteratively trains the search engine with feedback from all agents offline, then individually iteratively trains and serves the model for each agent during the online phase.}
    \label{fig:method}
    \vspace{-0.2cm}
\end{figure*}

\section{Problem Formulation}
\label{sec:problem}

Consider the retrieval model $R_\theta$, parameterized by $\theta$, whose main role is to facilitate information access from a corpus $C$ for a set of RAG models (a.k.a, RAG agents) denoted as $M = \{M_{i}\}_{i=1}^{n}$. Each $M_i$ acts as a black-box agent for $R_\theta$, which means that $R_\theta$ does not have access to the models' architecture, configuration, or parameters. Each $M_i$ is designed to perform a knowledge-intensive task $T_i = (D^{\text{train}}_i, D^{\text{test}}_i, \mu_i)$ that requires external information from the corpus $C$ as the knowledge source. There is a training dataset \( D^{\text{train}}_i = \{(x_j, y_j)\}_{j=1}^{|D^{\text{train}}_i|} \) for each agent \( M_i \), which can be used by the retrieval model $R_\theta$ for offline optimization. At inference, each agent \( M_i \) operates \textit{sequentially} on a test dataset \( D^{\text{test}}_i = \{(x'_j, y'_j)\}_{j=1}^{|D^{\text{test}}_i|} \) in the same order. The end-to-end performance of the agent $M_i$ can be measured by a utility function (metric) \( \mu_i \).

Each RAG agent can be simply formulated as $\bar{y}_{M_i} = M_i(x; R_\theta)$. In more detail, given an input $x$, each RAG agent $M_i$ submits a query to the retrieval model $R_\theta$ for information access, and generates the output by consuming the top $k$ retrieved documents (i.e., $\mathcal{R}_k=\{r_1, ..., r_k\}$). As suggested in \cite{urag}, the search engine and RAG agents can engage in an offline optimization process, in which each RAG agent $M_i$ produces a feedback list for a retrieval list of size $k$, denoted as $f_{M_i} = \{f_j\}_{j=1}^{k} \in \{0, 1\}^{1\times k}$. This feedback list indicates the usefulness of each retrieved document in the retrieval list from the agent's perspective.  Feedback can be computed in various ways; based on the performance of the generated output when utilizing the retrieved documents on the downstream task, e.g., using the utility function $\mu_i$, or based on ratings received from the users or evaluators of the RAG agent, among other methods. Without loss of generality, this paper focuses on the first case as the main method for providing feedback from the agents. In this approach, feedback is based on the agent's downstream performance when it uses each document individually in its task. Therefore, the main goal of this paper is to optimize the retrieval model $R_\theta$ to maximize the the expected utility for all the RAG agents that use the retrieval model.

\section{Learning to Rank with Iterative Utility Maximization}
\label{sec:method}

State-of-the-art commercial search engines have been trained using implicit feedback from their users, such as clicks, scrolling behavior, and dwell time \citep{10.1145/2568388.2568413, 10.1145/3018661.3018699, 4292009}. In our setting, however, the primary users of the search engine are the RAG agents that consume the retrieval results. Therefore, their feedback on the quality of the retrieval results serves as the primary signal for optimizing the search engine. Previous work \cite{urag} has explored the use of this feedback to train the model, where they employed the initial parameters of the search engine---without any prior training---to retrieve a set of documents and collect feedback for them in an offline setting to optimize the search engine. This approach has several shortcomings. First, using an untrained retrieval model to gather documents for feedback collection is suboptimal. If the parameters are not well-initialized, the quality of the initial document retrievals would be low, leading to a feedback collection process that may not yield relevant documents. Consequently, the model might easily learn to distinguish these poorly retrieved documents from relevant ones, resulting in an inadequately trained search engine. 

Furthermore, since the initial documents are retrieved without any prior feedback to adapt the search engine to each agent's information needs, they may not be relevant or useful to the agents. Additionally, relying solely on offline feedback collection limits the system, as it does not allow for continuous adaptation and improvement of the retrieval model based on real-time interactions. Implementing an online learning approach could address these issues by enabling the model to update its parameters incrementally during serving phase as new feedback is collected. This would allow the system to continuously refine its retrieval strategies, adjusting to the evolving information needs of each agent, improving personalization of the search engine for each agent.

We addresses the aforementioned issues by introducing the Iterative Utility Maximization (\ourframework) framework for training the retrieval model. This framework optimizes the retrieval model with the feedback collected from all agents during an offline phase, and it incorporates each agent's specific feedback during an online phase, all in an iterative manner to maximizes the probability of receiving positive feedback for each served query, as shown in Figure~\ref{fig:method}.

\subsection{Offline Ranking Optimization through Iterative Utility Maximization}
\label{sec:emrl-offline}

Similar to self-training of LLMs using self-generated data \citep{rest-em}, we define a \textit{optimality binary variable} $o$, where $p(o = 1|\mathcal{R}_k, x) \propto f_{M_i}(\mathcal{R}_k, x)$, meaning that variable $o=1$ indicates positive feedback and $o=0$ indicates negative feedback. Here, $\mathcal{R}_k$ is a retrieved list of $k$ documents, and $f_{M_i}(\mathcal{R}_k, x)$ represents the feedback for the retrieved list $\mathcal{R}_k$ given input $x$ from the RAG agent ${M_i}$. The main goal is to maximize log-likelihood of observing $o = 1$ for a given input $x$. Since in this process the retrieved documents affect the variable $o$, we can rewrite the log-likelihood as: $\log p(o = 1|x) = \log \sum_{\mathcal{R}_k \in \pi_k(C)} p_{\theta}(\mathcal{R}_k | x)p(o = 1 | x,\mathcal{R}_k)$, where $\pi_k(C)$ denotes all permutations of $k$ documents being selected from the corpus $C$. The summation over all possible retrieval lists $\mathcal{R}_k$ is computationally expensive for a large corpus,\footnote{In our case, the corpus contains approximately 36 million document chunks.} making the calculation infeasible. Instead of directly maximizing $\log p(o = 1|x)$, one can maximize its Evidence Lower Bound (\textit{ELBO}), denoted as $L(p_\theta, q)$, with respect to the retriever's parameters $\theta$ and a variational distribution $q(\mathcal{R}_k | x)$. The variational distribution is used to approximate the latent variable $\mathcal{R}_k$ distributions, simplifying and making the computation more efficient. Since  $L(p_\theta, q)$ is a lower bound for $\log p(o = 1|x)$, maximizing \textit{ELBO} ensures increase in $\log p(o = 1|x)$. Formally:
\begin{align}
    \log & p(o = 1|x) = \log \sum_{\mathcal{R}_k \in \pi_k(C)} p_{\theta}(\mathcal{R}_k | x)p(o = 1 | x,\mathcal{R}_k) \times \frac{q(\mathcal{R}_k|x)}{q(\mathcal{R}_k|x)} \nonumber \\ 
    &= \log \mathop{\mathbb{E}}_{\mathcal{R}_k \sim q(\mathcal{R}_k|x)}\left[ \frac{p(o = 1 | x, \mathcal{R}_k)p_\theta(\mathcal{R}_k|x)}{q(\mathcal{R}_k|x)} \right] \nonumber \\
    &\geq \mathop{\mathbb{E}}_{\mathcal{R}_k \sim q(\mathcal{R}_k|x)}\left[ \log \frac{p(o = 1 | x, \mathcal{R}_k)p_\theta(\mathcal{R}_k|x)}{q(\mathcal{R}_k|x)} \right] \quad \text{(Jensen's ineq.)} \nonumber \\ 
    &= -\KL[q(\mathcal{R}_k|x) || p(o = 1 | x, \mathcal{R}_k)p_\theta(\mathcal{R}_k|x)] \nonumber \\ 
    &= \mathop{\mathbb{E}}_{\mathcal{R}_k \sim q(\mathcal{R}_k|x)}\left[ \log p(o = 1 | x, \mathcal{R}_k) \right] -\KL[q(\mathcal{R}_k|x) || p_\theta(\mathcal{R}_k|x)]  \nonumber \\ 
    &\eqcolon L(p_\theta, q)
    \label{eq:after-jensen}
\end{align}
where $\KL$ denotes the Kullback-Leibler divergence between the two given distributions. Thus, we show that $\log p(o = 1|x) \geq L(p_\theta, q)$, which means that maximizing $L(p_\theta, q)$ results in increasing the lower bound of $\log p(o = 1|x)$. To maximize $L(p_\theta, q)$, we utilize an Expectation-Maximization (EM) algorithm as follows.

\subsubsection*{\textbf{Expectation Step ($q^{t+1} = \operatorname*{arg\,max}_q L(p_{\theta^{t}}, q)$):}} 
Maximizing $q$ is considered the Expectation step since it involves finding the distribution 
$q$ that best approximates the true posterior distribution of the latent variable $\mathcal{R}_k$. Considering the formulation in Equation \ref{eq:after-jensen}, where $L(p_\theta, q) = -\KL[q(\mathcal{R}_k|x) || p(o = 1 | x, \mathcal{R}_k)p_{\theta^t}(\mathcal{R}_k|x)]$, it implies that maximum of $L(p_\theta, q)$ occurs when $q^{t+1} = p(o = 1 | x, \mathcal{R}_k)p_{\theta^t}(\mathcal{R}_k|x)$, because the KL divergence is non-negative and equals to zero when the given two distributions are identical.

\subsubsection*{\textbf{Maximization Step ($\theta^{t+1} = \operatorname*{arg\,max}_\theta L(p_\theta, q^{t+1})$):}} 

The goal of this step is to update the model parameters by maximizing the expected log-likelihood computed in the Expectation step, thereby fit the model to the observed data. We re-write this step as:
\begin{align}    
    \theta^{t+1} & = \operatorname*{arg\,max}_\theta L(p_\theta, q^{t+1}) = \operatorname*{arg\,max}_\theta - \KL[q^{t+1} || p_{\theta}(\mathcal{R}_k|x)] \nonumber \\ 
    & = \operatorname*{arg\,max}_\theta \sum_{\mathcal{R}_k} q^{t+1} \log p_{\theta}(\mathcal{R}_k|x) \nonumber \\
    & = \operatorname*{arg\,max}_\theta \sum_{\mathcal{R}_k} p(o = 1 | x, \mathcal{R}_k)p_{\theta^t}(\mathcal{R}_k|x) \log p_{\theta}(\mathcal{R}_k|x) \nonumber \\ 
    &= \operatorname*{arg\,max}_\theta \mathop{\mathbb{E}}_{\mathcal{R}_k \sim p_{\theta^t}(\mathcal{R}_k|x)} \left[ p(o = 1 | x, \mathcal{R}_k)\log p_{\theta}(\mathcal{R}_k|x) \right]
\end{align}

Since feedback in our case is non-negative, as assumed in Section \ref{sec:problem}, and $p(o = 1 | x, \mathcal{R}_k) \propto f_{M_i}(\mathcal{R}_k, x)$, the final objective function considering all agents to get updated parameters $\theta^{t+1}$ is defined as:
\begin{align}
    \label{eq:offline-rl}
    \operatorname*{arg\,max}_\theta   \mathop{\mathbb{E}}_{M_i \sim M} \left[ \mathop{\mathbb{E}}_{x \sim D^{\text{train}}_i} \left[ \mathop{\mathbb{E}}_{\mathcal{R}_k \sim p_{\theta^t}(\mathcal{R}_k|x)} \left[ f_{M_i}(\mathcal{R}_k, x) \log p_\theta(\mathcal{R}_k|x) \right] \right] \right] 
\end{align}
where the retrieval lists are sampled from the parameters of the retrieval model in the previous iteration ($\mathcal{R}_k \sim p_{\theta^t}(\mathcal{R}_k|x)$). As noted by \citet{erag}, gathering agent's feedback for an entire ranked list is computationally expensive, and collecting feedback on a per-document basis is more efficient. Additionally, \citet{NEURIPS2021_da3fde15} emphasize the difficulty of calculating a retrieval list probabilities. To simplify this, we instead use a pointwise ranking objective function \cite{pointwise} using the same technique. 

To implement this approach, two adjustments are needed: First, for pointwise learning-to-rank, sampling should be based on the probabilities of the documents, which can be approximated by their relevance to the query. Let $p_{\theta^t}(R=1|x, d)$ represent the probability that document $d$ is relevant to query $x$. We assume that the more relevant a document is to a query, the higher its probability. Therefore, we have: $p^*_{\theta^t}(d|x) \propto p_{\theta^t}(R=1|x, d)$. Additionally, we assume that any documents outside the top $k$, ranked by $p_{\theta^t}(R=1|x, d)$, have a probability of zero. Thus, sampling from the retrieval list $\mathcal{R}k$ can be approximated by selecting the top $k$ documents based on $p^*_{\theta^t}(d|x) \propto p_{\theta^t}(R=1|x, d)$. The second adjustment is that the objective function should aim to maximize the probability of documents that receive positive feedback. Thus, we can reformulate the problem by using the feedback as a relevance label, training the model to classify whether a document is relevant or not. We can define the final objective to get updated parameters $\theta^{t+1}$ as: 
\begin{align}
    \label{eq:offline-rl-ours}
    \operatorname*{arg\,max}_\theta  \mathop{\mathbb{E}}_{M_i \sim M} \left[ \mathop{\mathbb{E}}_{x \sim D^{\text{train}}_i} \left[ \mathop{\mathbb{E}}_{d \sim p^*_{\theta^t}(d|x)} \left[ \log p_\theta(R = f_{M_i}(d, x)|x, d) \right] \right] \right]
\end{align}

\subsubsection*{\textbf{Training Procedure:}} 

To train the retrieval model $R_\theta$, we assume $T$ expectation and maximization iterations are performed. In each iteration, the RAG agents submit their training queries to the retrieval model, which then provides them with the top results based on the given query from the corpus $C$. In response, the agents return feedback per retrieved document. After collecting feedback from all agents, the search engine enters an optimization phase where Equation \ref{eq:offline-rl-ours} is used to optimize the model based on the provided feedback. This procedure is shown in Algorithm \ref{alg:exp-max-offline-rl}.

\begin{algorithm}[H]
\caption{The \rlem algorithm.}\label{alg:exp-max-offline-rl}
\begin{algorithmic}
    \STATE \textbf{Initialize} the retrieval model from a pretrained encoder checkpoint
    \FOR{$t = 1$ to $T$}
        \STATE $D^t = \{\}$
        \FOR{$M_i$ in $M$}
            \FOR{$x$ in $D_i$}
                \STATE \textbf{Retrieve} $k$ docs from corpus $C$ by $R_{\theta^{t}}$: $r = R_{\theta^{t}}(x, C, k)$
                \STATE \textbf{Collect} feedback of agent $M_i$ for list $r$: $f_i = f_{M_i}(d_i, x)$
                \STATE \textbf{Add} feedback to iteration $t$ dataset: $D^t = D^t \cup \{(x, d_i, f_i)\}$
            \ENDFOR
        \ENDFOR
        \STATE ${\theta^{t+1}} = \operatorname*{arg\,max}_\theta \mathop{\mathbb{E}}_{(x, d, f) \sim D^t} \left[ \log p_\theta(R = f|x, d) \right]$
    \ENDFOR
\end{algorithmic}
\end{algorithm}

\subsection{Online Ranking Optimization through Iterative Utility Maximization}

In the previous subsection, $R_\theta$ collects feedback from all RAG agents on their respective training sets in an offline, iterative fashion, and uses their feedback for optimizing its parameters. However, this approach is not feasible during the serving phase\footnote{The serving phase refers to the period when the system is being tested.} for several reasons. First, rapid access to information is necessary for the serving phase, since agents cannot afford to wait for feedback to be collected from all other agents. Additionally, offline optimization methods typically require access to the complete set of queries for effective optimization, which is not applicable in this context, where inputs appear sequentially. Therefore, this phase must rely on online optimization of the search engine tailored for each individual agent, utilizing their specific feedback. For this purpose, we assume that feedback for a test instance $x_t$ from the agent $M_i$ is only provided after $R_\theta$ has already retrieved and delivered the relevant documents to the agent, and the agent will not submit $x_t$ to the system again afterwards. In other words, the search engine has only observed and served the first $t-1$ queries and received feedback for them by the time it attempts to serve $x_t$. Therefore, the search engine can utilize the feedback from the first $t-1$ queries to optimize its performance for serving the $t$\textsuperscript{th} query.

There are several strategies for updating the search engine parameters based on received online feedback. One possible approach is to update the search engine after each feedback to align it with the agent's preferences. However, this method has several shortcomings: first, updating the search engine after each feedback is computationally expensive. Additionally, the feedback for a single input may be noisy, leading to noisy gradient updates. On the other hand, an alternative strategy is to observe the majority of test instances before updating the model for the remaining queries. However, this would result in using the same old parameters for most of the test instances. Therefore, we adopt a middle-ground approach and apply the same method as described in Section \ref{sec:emrl-offline} to optimize the model over several iterations. We define an iteration as serving $b$\footnote{We call this parameter the online optimization batch size, which differs from the batch size used in gradient optimization methods.} consecutive queries and receiving the corresponding feedback. Thus, in step $t+1$, the model has access to the previous user queries $Q^{t}_{M_i} = \{x_1, ..., x_{b\times t}\}$ (This is the \textbf{Expectation Step}, as defined in Section \ref{sec:emrl-offline}). Finally, the parameters for agent $M_i$ can be updated based on the following objective function (This is the \textbf{Maximization Step}, as defined in Section \ref{sec:emrl-offline}):
\begin{equation}
    \label{eq:in-session-learning}
    \theta^{t+1}_{M_i} = \operatorname*{arg\,max}_\theta \mathop{\mathbb{E}}_{x \sim Q^{t}_{M_i}} \left[ \mathop{\mathbb{E}}_{d \sim p^*_{\theta^t_{M_i}}(d|x)} \left[ \log p_\theta(R = f_{M_i}(d, x)|x, d) \right] \right]
\end{equation}
where the main difference between this objective function and Equation \ref{eq:offline-rl} is that the optimization occurs for each individual agent based on their own feedback, with inputs sampled from the previous queries observed from the agent during the serving phase. 

\subsubsection*{\textbf{Training Procedure:}} 
To optimize the search engine using \sessionlearning, we assume a specific agent submits $b$ queries to the search engine, which retrieves the top results from the corpus $C$ for the given queries and the previous iteration parameters. In response, the agent provides feedback for the retrieved documents. Following this, the search engine enters an optimization phase, updating its parameters based on the feedback received for all queries submitted thus far by that agent, using Equation \ref{eq:in-session-learning}, to get the parameters for next iteration. This procedure is shown in Algorithm \ref{alg:in-session-learning}. 

\begin{algorithm}[H]
\caption{The \sessionlearning algorithm.}\label{alg:in-session-learning}
\begin{algorithmic}
    \STATE \textbf{Initialize} $t = 1$
    \STATE \textbf{Initialize} the retrieval model for agent $M_i$ from Algorithm \ref{alg:exp-max-offline-rl} checkpoint
    \STATE $D^{test}_{M_i} = \{\}$
    \WHILE{there is a query $x_i$}
        \STATE \textbf{Retrieve} $k$ docs from corpus $C$ by $R_{\theta^{t}}$: $r = R_{\theta^{t}_{M_i}}(x_i, C, k)$
        \STATE \textbf{Collect} feedback of agent $M_i$ for list $r$: $f_j = f_{M_i}(d_j, x_i)$
        \STATE \textbf{Add} feedback to online dataset: $D^{test}_{M_i} = D^{test}_{M_i} \cup \{(x_i, d_j, f_j)\}$
        \IF{$|D^{test}_{M_i}|$ dividable by $b$}
            \STATE ${\theta^{t+1}_{M_i}} = \operatorname*{arg\,max}_\theta \mathop{\mathbb{E}}_{(x, d, f) \sim D^{test}_{M_i}} \left[ \log p_{\theta}(R = f|x, d) \right]$
            \STATE $t = t+1$
        \ENDIF
    \ENDWHILE
\end{algorithmic}
\end{algorithm}

\subsection{The Search Engine Architecture}
The proposed algorithms can work for any learning-to-rank approach.
Following uRAG from \citet{urag}, this paper employs a two-stage cascaded retrieval system. In the first stage, BM25 \cite{bm25} is used to retrieve a set of initial relevant candidate documents. Then, a fine-tuned cross-encoder model is applied to re-rank the retrieved documents. Note that all the trainable parameters $\theta$ in the search engine are for the re-ranking model. Following \citet{bert-reranker}, we add a linear projection over the representation of the start token (i.e., [CLS]) of a text encoder to obtain the relevance probability of a query and document, as follows:
\begin{equation}
    p(R=1|x, d) = \sigma(\text{ENCODER}(tid; mid; x; d)\cdot W)
\end{equation}
where $d$ is a document, $x$ is the query, $tid$ is an ID associated with the agent's task, $mid$ is the ID associated with the LLM backbone used in the agent, $\sigma$ is the sigmoid function, and $W \in \mathbb{R}^{D\times1}$ is a linear projection, where $D$ represents the embedding dimensionality of the encoder. We utilize BERT\footnote{Checkpoint can be find at: \url{https://huggingface.co/google-bert/bert-base-uncased}} \cite{bert} with 110M parameters and embedding dimensionality of $D=768$ as the encoder. Here, following \citet{urag}, $tid$ and $mid$ is used for the purpose of personalizing retrieval model for each agent. In this way, each agent provides its underlying LLM architecture and the task it is performing, represented by a task ID ($tid$) and a model ID ($mid$). The search engine uses these identifiers as input and, when optimized with feedback from each user, learns both the specific preferences of individual users (presence of $tid$ and $mid$ together in input) and the general information needs associated with each task ($tid$) and architecture ($mid$). This enables the search engine to better understand user preferences and provide more effective, tailored results while also learning the general preferences associated with each task and agent architecture.

\section{Experiments}

\subsection{Experiment Setup}

\subsubsection*{\textbf{Datasets \& Corpus}}
We use various tasks from the KILT benchmark \citep{kilt} to evaluate our approach. 
Dataset statistics is reported Table~\ref{tab:data}. 
We experiment with six diverse datasets, including three diverse open-domain question answering datasets: Natural Questions (NQ) \citep{kwiatkowski-etal-2019-natural}, TriviaQA \citep{joshi-etal-2017-triviaqa}, and HotPotQA \citep{yang-etal-2018-hotpotqa}. Notably, HotPotQA focuses on questions requiring multi-hop reasoning. Additionally, we use one fact verification dataset, FEVER \citep{thorne-etal-2018-fever}, and two slot-filling datasets for relation extraction: zsRE \citep{levy-etal-2017-zero} and T-REx \citep{elsahar-etal-2018-rex}. Note that since the T-REx training set contains approximately 2.2 million samples, we randomly select 5\% of them to train our models in order to speed up the experiments. We use the same samples as is used in \cite{urag}. It is also important to mention that the test set labels for these datasets are not publicly available. Therefore, we directly use the validation set to evaluate the models. Note that we utilize the McNemar statistical significance test \citep{McNemar1947} in our experiments, as the metrics for the tasks—accuracy and exact match—yield binary outcomes, making this test appropriate for evaluating significant performance differences. We use the Wikipedia dump provided by the KILT benchmark as the unstructured knowledge source.\footnote{The retrieval corpus is available at \url{https://dl.fbaipublicfiles.com/ur/wikipedia_split/psgs_w100.tsv.gz}} We follow the pre-processing method outlined by \citet{dpr}, in which each document is split into passages with a maximum length of 100 words. Additionally, the document title is concatenated with each passage to form the entries in the retrieval corpus.

\subsubsection*{\textbf{Agents Configuration}}

\begin{table}
    \centering
    \caption{A list of RAG models used in our experiments for training and evaluation.}
    \label{tab:predictive_models}
    \vspace{-0.4cm}
    \adjustbox{width=\linewidth}{
    \begin{tabular}{l|llllc}
         & \textbf{Task} & \textbf{Data} & \textbf{Utility Func.} & \textbf{LM} & \textbf{\#Docs} \\\hline
        $M_1$ & open-domain QA & NQ & Exact Match & RA-T5 & 10 \\
        $M_2$ & open-domain QA & NQ & Exact Match & RA-BART & 4 \\
        $M_3$ & open-domain QA & NQ & Exact Match & FiD & 10 \\\hdashline
        $M_4$ & open-domain QA & TriviaQA & Exact Match & RA-T5 & 10 \\
        $M_5$ & open-domain QA & TriviaQA & Exact Match & RA-BART & 4 \\
        $M_6$ & open-domain QA & TriviaQA & Exact Match & FiD & 10 \\\hdashline
        $M_7$ & open-domain QA & HotPotQA & Exact Match & RA-T5 & 10 \\
        $M_8$ & open-domain QA & HotPotQA & Exact Match & RA-BART & 4 \\
        $M_9$ & open-domain QA & HotPotQA & Exact Match & FiD & 10 \\\hline
        $M_{10}$ & fact verification & FEVER & Accuracy & RA-T5 & 10 \\
        $M_{11}$ & fact verification & FEVER & Accuracy & RA-BART & 4 \\
        $M_{12}$ & fact verification & FEVER & Accuracy & FiD & 10 \\\hline
        $M_{13}$ & slot filling & zsRE & Accuracy & RA-T5 & 10 \\
        $M_{14}$ & slot filling & zsRE & Accuracy & RA-BART & 4 \\
        $M_{15}$ & slot filling & zsRE & Accuracy & FiD & 10 \\\hdashline
        $M_{16}$ & slot filling & T-REx & Accuracy & RA-T5 & 10 \\
        $M_{17}$ & slot filling & T-REx & Accuracy & RA-BART & 4 \\
        $M_{18}$ & slot filling & T-REx & Accuracy & FiD & 10 \\\hline
    \end{tabular}}
\end{table}

\begin{table}[t]
    \centering
    \caption{A list of datasets from KILT \cite{kilt} used in our experiments. The validation data from KILT is used as test sets. $^*$ Given the large training set in the original T-REx dataset, we only sampled 5\% of data for training our models.}
    \vspace{-0.4cm}
    \begin{tabular}{ll|ll}
        & \textbf{Dataset} & \textbf{\#train} & \textbf{\#test} \\ \hline
        \multicolumn{2}{l|}{\textbf{open-domain QA (short answer)}} & \\
        & Natural Questions & 87,372 & 2,837 \\
        & TriviaQA & 61,844 & 5,359 \\
        & HotPotQA & 88,869 & 5,600 \\\hdashline
        \multicolumn{2}{l|}{\textbf{fact verification}} & \\
        & FEVER & 104,966 & 10,444 \\\hdashline
        \multicolumn{2}{l|}{\textbf{slot-filling relation extraction}} & \\
        & zsRE & 147,909 & 3,724 \\
        & T-REx & 114,208$^*$ & 5,000 \\\hline
    \end{tabular}
    \label{tab:data}
\end{table}

\begin{table*}[t]
    \centering
    \caption{Downstream performance of RAG models in Table~\ref{tab:predictive_models} utilizing the search engine. Superscripts $1$, $2$, $3$, $4$, and $5$ denote statistically significant improvements in the performance compared to BM25, Contriever, $\text{Reranker}_{\text{individual}}$, $\text{Reranker}_{\text{dataset}}$, uRAG, respectively, using McNemar significance test ($p<0.05$).}
    \vspace{-0.4cm}
    \resizebox{\linewidth}{!}{
    \begin{tabular}{ll|cc|lll|ll}
        \textbf{RAG} & \multirow{2}{*}{\textbf{Data \& Metric}} & \multirow{2}{*}{\textbf{BM25}} & \multirow{2}{*}{\textbf{Contriever}} & \multirow{2}{*}{\textbf{$\text{Reranker}_{\text{individual}}$}}& \multirow{2}{*}{\textbf{$\text{Reranker}_{\text{dataset}}$}} & \multirow{2}{*}{\textbf{uRAG}} & \multicolumn{2}{c}{\textbf{\ourframework}} \\
        \textbf{Model} & & & & & & & \rlem & \sessionlearning \\\hline
        $M_1$ & NQ - EM & 28.05 & 22.55 & 36.76 & 37.39 & 37.82 & {38.42}$^{12}$ & \textbf{38.84}$^{1234}$ \\
        $M_2$ & NQ - EM & 33.09 & 23.68 & 40.07 & 40.50 & 42.12 & {43.60}$^{12345}$ & \textbf{45.11}$^{12345}$ \\
        $M_3$ & NQ - EM & 29.64 & 23.69 & 40.50 & 41.14 & 42.37 & \textbf{42.51}$^{1234}$ & 42.47$^{1234}$ \\\hdashline
        $M_4$ & TriviaQA - EM & 51.35 & 44.33 & 59.28 & 60.25 & 60.68 & \textbf{61.87}$^{12345}$ & 61.59$^{12345}$ \\
        $M_5$ & TriviaQA - EM & 57.52 & 48.49 & 64.76 & 67.23 & {68.12} & 68.03$^{123}$ & \textbf{68.87}$^{1234}$ \\
        $M_6$ & TriviaQA - EM & 60.48 & 49.30 & 67.44 & 68.63 & 68.74 & \textbf{70.27}$^{12345}$ & 70.05$^{12345}$ \\\hdashline
        $M_7$ & HotPotQA - EM & 27.51 & 18.80 & 29.92 & 30.91 & 31.33 & \textbf{31.41}$^{123}$ & 31.16$^{123}$ \\
        $M_8$ & HotPotQA - EM & 31.21 & 20.78 & 35.03 & 34.62 & 34.85 & \textbf{35.51}$^{124}$ & 35.41$^{12}$ \\
        $M_9$ & HotPotQA - EM & 29.48 & 20.43 & 32.54 & 32.71 & 33.46 & \textbf{33.77}$^{1234}$ & 33.75$^{1234}$ \\\hline
        $M_{10}$ & FEVER - Accuracy & \textbf{86.83} & 84.21 & 86.24 & \textbf{86.83} & 86.46 & 86.58$^{2}$ & 86.58$^{2}$ \\
        $M_{11}$ & FEVER - Accuracy & \textbf{87.54} & 84.37 & 84.38 & \textbf{87.54} & 85.99 & 86.17$^{23}$ & {87.44}$^{235}$ \\
        $M_{12}$ & FEVER - Accuracy & {87.04} & 86.74 & 86.02 & {87.04} & 86.55 & 86.98$^{235}$ & \textbf{87.47}$^{235}$ \\\hline
        $M_{13}$ & zsRE - Accuracy & 55.37 & 38.77 & 60.39 & 59.98 & 61.09 & {61.22}$^{1234}$ & \textbf{61.89}$^{12345}$ \\
        $M_{14}$ & zsRE - Accuracy & 51.42 & 29.05 & 59.29 & 58.96 & 60.58 & \textbf{60.68}$^{1234}$ & 60.31$^{1234}$ \\
        $M_{15}$ & zsRE - Accuracy & 55.42 & 37.35 & 60.47 & 60.66 & 62.13 & {62.35}$^{1234}$ & \textbf{62.43}$^{1234}$ \\\hdashline
        $M_{16}$ & T-REx - Accuracy & 70.88 & 56.94 & 73.58 & 72.86 & 72.92 & {73.62}$^{12345}$ & \textbf{73.98}$^{12345}$ \\
        $M_{17}$ & T-REx - Accuracy & 75.16 & 58.30 & 80.04 & 80.18 & 79.94 & \textbf{80.32}$^{12}$ & 80.24$^{12}$ \\
        $M_{18}$ & T-REx - Accuracy & 78.88 & 65.06 & 80.78 & 80.34 & 80.24 & {80.88}$^{125}$ & \textbf{81.14}$^{12345}$ \\\hline
        Overall & (macro-average) & 55.38 & 45.15 & 59.86 & 60.43 & 60.85 & {61.34}$^{12345}$ & \textbf{61.59}$^{12345}$ \\\hline
    \end{tabular}
    }
    \label{tab:main-results}
\end{table*}

Following \citet{urag}, we use 18 diverse RAG agents, as listed in Table \ref{tab:predictive_models}. In this setting, each agent is trained on a separate dataset, with a distinct set of resources, a different underlying LLM, and retrieves a varying number of documents to perform its task. Each RAG agent is fine-tuned on the corresponding training set. Additionally, the number of retrieved documents is determined based on each agent's LLM maximum input size. We consider two types of RAG agents: 
\begin{enumerate}[leftmargin=*]
    \item Retrieval-augmented LLM (RA-X) is a language model that consumes $k$ documents per input via in-prompt augmentation based on the following input format: ``\texttt{\{input\} context 1: \{doc1\} \ldots  context k: \{dock\}}'', where \texttt{\{input\}} is $x$ and \texttt{\{doci\}} denotes the content of the $i$\textsuperscript{th} retrieved document.

    \item Fusion-in-Decoder (FiD) \citep{fid} uses a different augmentation approach. Unlike RA-X that is based on in-prompt augmentation, FiD first encodes the input and each retrieved document separately and uses the concatenation of all document encodings as cross-attention for the decoder. FiD can thus only be done using encoder-decoder language models.
\end{enumerate}

We utilize T5-small~\citep{t5} with 60M parameters and BART-base~\citep{bart} with 140M parameters using the first retrieval-augmentation approach and T5-small with the FiD. We use six datasets and apply three distinct RAG models to each, resulting in 18 unique agents for our experiments. All RAG agents are fine-tuned individually. The AdamW optimizer with a weight decay of $10^{-2}$ and a learning rate of $5 \times 10^{-5}$ for 10 epochs is used to train these models. A linear warmup is applied to the first 5\% of training steps. The effective batch size is set to 64 through gradient accumulation. Each model is trained on varying computational resources, including up to 8 A100, 1080ti, and 2080ti Nvidia GPUs. To optimize the RAG agents, in this paper, we employ a sequence-two-sequence loss function \cite{seq2seq} using cross-entropy between the predicted token probability and the ground truth output sequence.

Each RAG agent is expected to provide feedback for the given retrieval list as a singal to help improving the search engine. For simplicity, in this paper, we consider the case where the RAG model provides feedback on a per-document basis. To generate feedback, we use the evaluation metric associated with the dataset that the model is applied to as the utility function for that model. Table~\ref{tab:predictive_models} outlines the utility functions assigned to the RAG agents used in this study. Specifically, we employ Exact Match (EM) for question answering datasets and accuracy for the remaining tasks. For EM, we follow the post-processing procedures introduced by \citet{squad}. Accordingly, the feedback for a specific retrieval list reflects the performance of the RAG agent when using only each single document in the list in its task, and is evaluated based on the utility function and downstream task expected output.

\subsubsection*{\textbf{Search Engine Configuration}}

The search engine initially retrieves 100 documents using BM25 \citep{bm25}. These documents are then reranked using the appropriate reranking model based on the experimental setup. Training $R_\theta$ involves two phase. For the first phase of training with \rlem, the Adam optimizer \citep{adam} with a learning rate of $10^{-5}$ for two epochs is used. A linear warmup is applied to the first 5\% of training steps. The effective batch size is set to 512, achieved through gradient accumulation. A consistent value of $k = 32$ is used during training, and the maximum input length for this model is set to 256 tokens. We train the model for $T = 3$ iterations. Note that we randomly replace the model ID and task ID with the \textit{``unk''} token for 10\% of the training samples for better generalization. For the second phase training using \sessionlearning, we set batch size $b  = 256$. Additionally, we utilize the same optimizer and learning rate to train the model for two epochs in each iteration. In this case, we set the variable $k$ to be consistent with the downstream RAG agent configuration for each model in Table~\ref{tab:predictive_models}, in contrast with the offline phase, where we set a constant number of $k=32$ for all agents. Additionally, we do not replace the model ID and task ID with the \textit{``unk''} token for the sake of fully personalizing the search engine for the agent. 

\begin{figure}
    \centering
    \subfloat[\centering The Effect of Offline Learning Number of Iterations]{\includegraphics[width=0.43\linewidth]{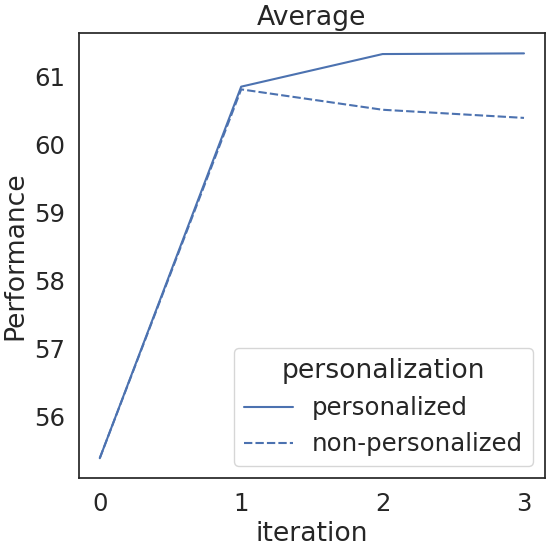}}
    \qquad
    \subfloat[\centering The Effect of Online Learning Batch Size]{\includegraphics[width=0.43\linewidth]{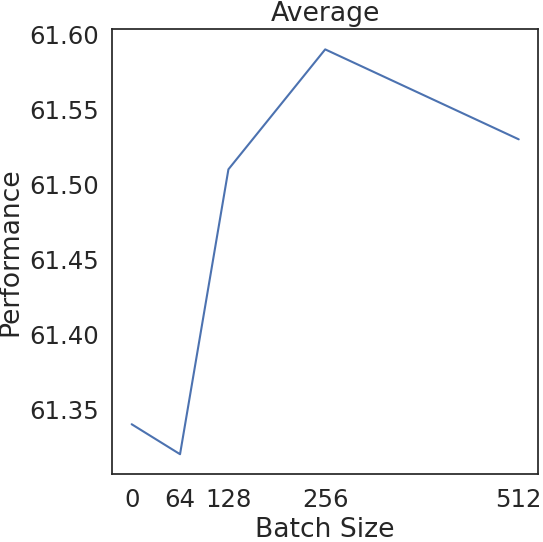}}
    \vspace{-0.4cm}
    \caption{(a) The impact of iterations on personalized (solid) and non-personalized (dashed) \rlem and (b) the effect of batch size $b$ on \sessionlearning performance.}
    \vspace{-0.4cm}
    \label{fig:iteration-perfromance}
\end{figure}

\subsection{Main Results}

This section addresses the following research questions, providing detailed analyses and insights based on our experimental findings.

\subsubsection*{\textbf{How does \rlem impact downstream RAG performance?}}

We employ the baselines introduced by \citet{urag} for comparative evaluation against our proposed approach. There are two types of baselines: (1) Single-stage retrieval and (2) Cascaded two-stage retrieval models. The single-stage retrieval baselines include BM25 \citep{bm25} as a sparse retriever and Contriever \citep{contriever} as a dense retriever. The cascaded two-stage methods employ BM25 in the first stage, followed by a reranker. \citet{urag} introduces three baselines for reranking: 1) $\text{Reranker}_{\text{individual}}$, which is a reranker trained individually for each agent based on a single round of feedback from the respective agent, 2) $\text{Reranker}_{\text{dataset}}$ that is a reranker trained for each dataset similar to previous one, using feedback from all agents associated with that dataset, and 3) uRAG, which is a single reranker trained across all agents using the combined feedback from different agents. The results of this experiment are presented in Table~\ref{tab:main-results} and suggest that \rlem outperforms all baselines for 14 out of 18 RAG agents, with the performance difference being statistically significant in 4 cases from all baselines. Moreover, \rlem achieves statistically significant improvements over all baselines when considering the average performance across all agents. That said, the results from this experiment suggest that \rlem is a promising and effective approach for addressing the problem at hand.


\subsubsection*{\textbf{How does the number of training iterations in \rlem impact the performance?}} 

An important hyperparameter for \rlem is the number of iterations. To investigate its impact, we train the search engine over several iterations and evaluate its performance at each step. The outcomes of this experiment are presented in Figure~\ref{fig:iteration-perfromance} (a---solid lines) for the average performance and Figure~\ref{fig:iteration-perfromance-tasks} for per agent performance. The results in Figure~\ref{fig:iteration-perfromance} (a) suggest that, overall, increasing the number of iterations leads to improved performance for \rlem. However, the magnitude of improvement diminishes with each additional iteration. For example, the improvement in average performance from iteration 0 to iteration 1 is 9\%, while the gain from iteration 1 to 2 is 0.8\%, and from iteration 2 to 3, the increase is less than 0.01\%. This indicates diminishing returns as the number of iterations increases.



\begin{figure}
    \centering
    \includegraphics[width=\linewidth]{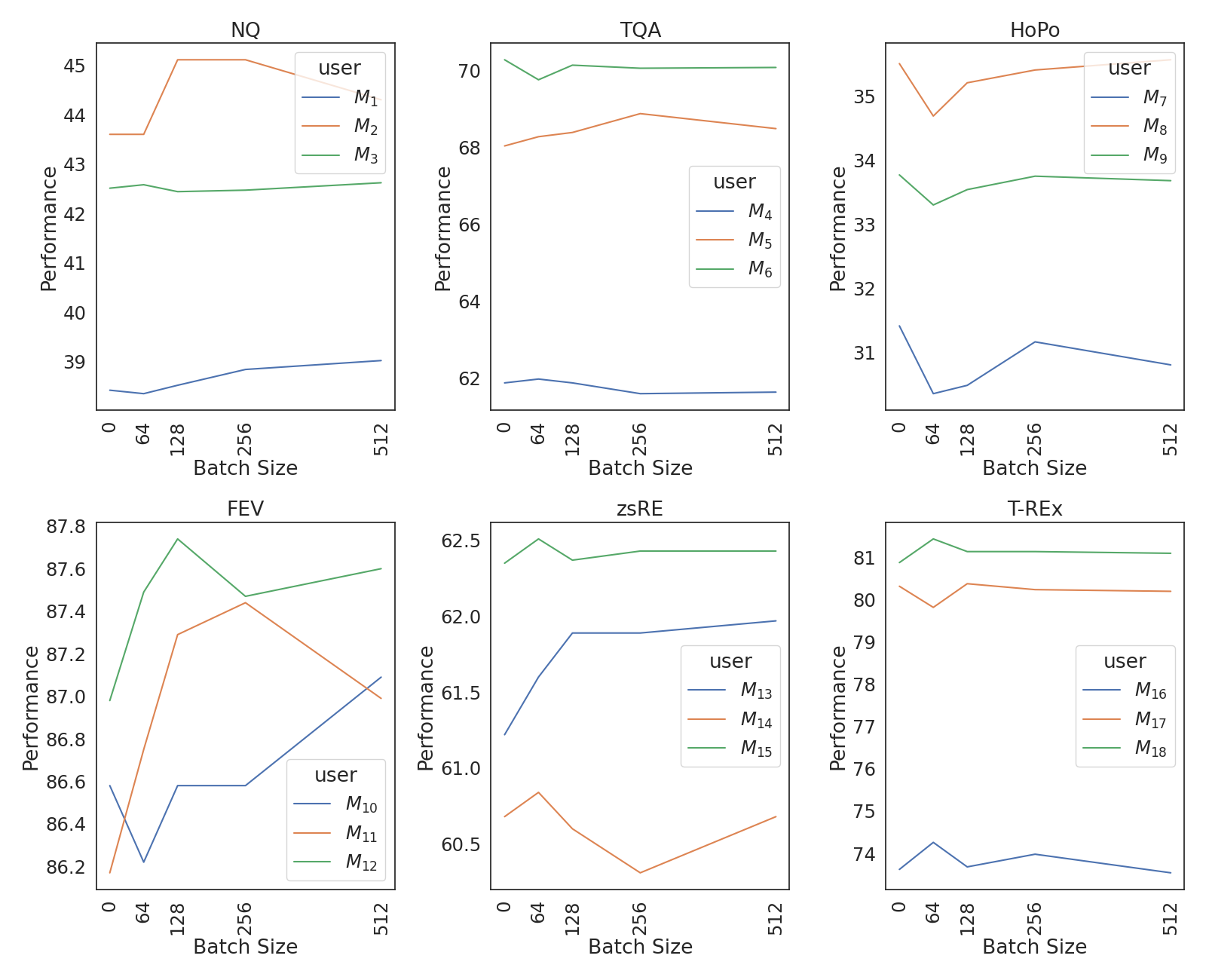}
    \vspace{-0.6cm}
    \caption{Effect of online learning batch size $b$ on \sessionlearning on per agent performance.}
    \label{fig:online-learning-batch-tasks}
\end{figure}

\subsubsection*{\textbf{How does ``personalization'' in \rlem training phase impact the performance?}}

\begin{figure}
    \centering
    \includegraphics[width=\linewidth]{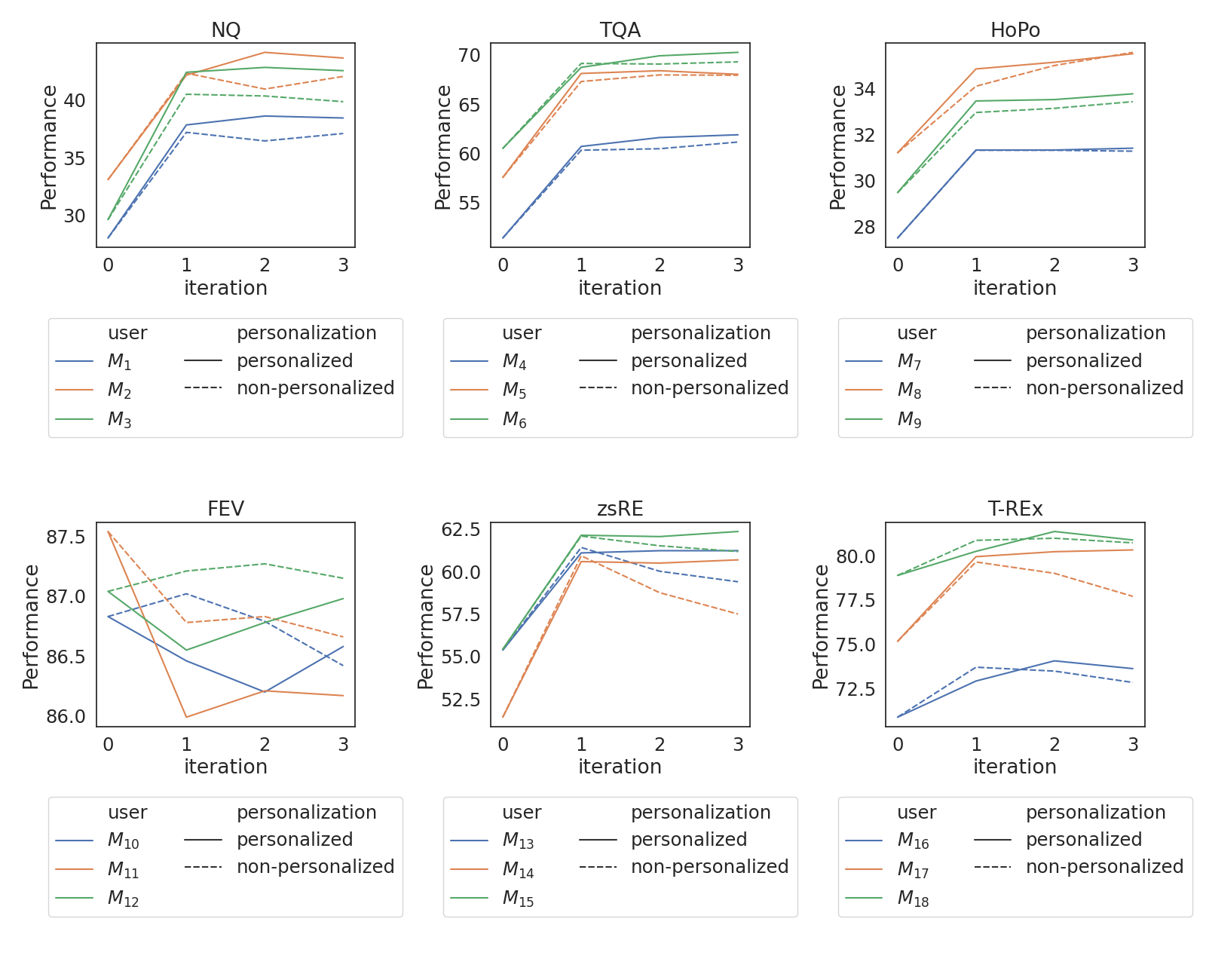}
    \vspace{-0.8cm}
    \caption{Impact of iteration count on personalized (solid) and non-personalized (dashed) \rlem plotted per agent.}
    \label{fig:iteration-perfromance-tasks}
    \vspace{-0.4cm}
\end{figure}

As explained in Section \ref{sec:method}, we use task ID (\textit{tid}) and model ID (\textit{mid}) to \textit{personalize} the search engine for each agent during the \rlem training. In this experiment, we remove these IDs and train the search engine without them, using the same setup as \rlem. This allows us to examine the impact of these identifiers on personalization of the search engine during the \rlem training process on the overall performance. The results of this experiment are demonstrated in in Figure~\ref{fig:iteration-perfromance} (a) for average performance (dashed lines) and Figure~\ref{fig:iteration-perfromance-tasks} for per agent performance (dashed lines).

The results in Figure~\ref{fig:iteration-perfromance-tasks} indicate that training the search engine without incorporating task and model IDs leads to a decline in the performance with increasing iterations. Specifically, while most agents benefit from the initial training round, performance get worse for 10 out of 18 agents in the second iteration. Additionally, in Figure~\ref{fig:iteration-perfromance} (a), the average performance of the search engine also decreases with additional iterations. The results show that the performance gap between the personalized and non-personalized search engines widens with more iterations. Specifically, the plot reveals a growing divergence between the average performance of the personalized search engine and the non-personalized one as the number of iterations increases. This suggests that incorporating personalization significantly benefits the search engine, with the performance gains becoming more pronounced over time. We believe this phenomenon is likely due to the lack of personalization in the non-personalized search engine. Without task and model IDs, the search engine retrieves more generic documents in subsequent iterations, which may not be particularly useful for specific agents. Consequently, the search engine may begin to overfit to the average preferences of all agents during the offline phase, rather than effectively addressing the unique needs of each individual agent.

\subsubsection*{\textbf{How does \sessionlearning impact downstream RAG performance?}}

\begin{figure*}
    \centering
    \includegraphics[width=\linewidth]{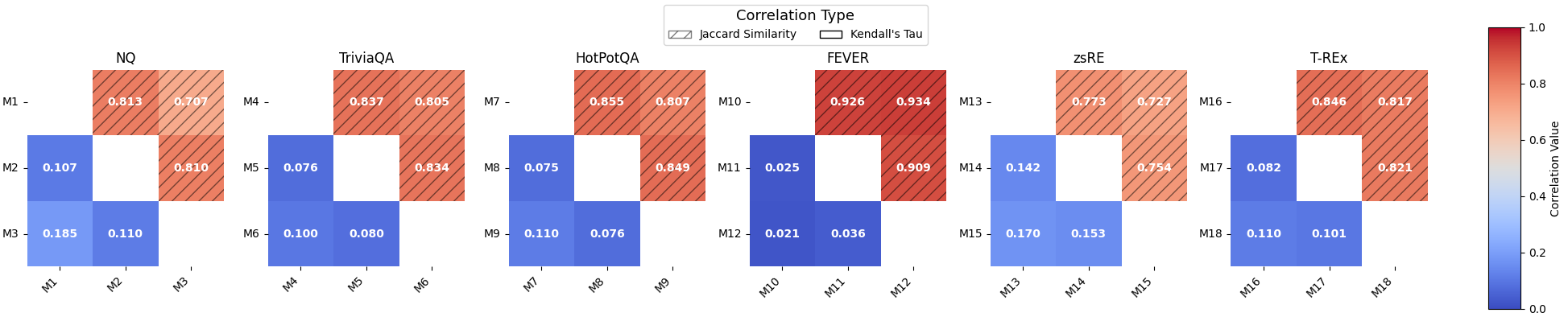}
    \vspace{-0.8cm}
    \caption{The Kendall's tau and Jaccard's similarity between the retrieval lists for agents that utilize the same dataset.}
    \label{fig:ret-corr}
\end{figure*}

To investigate this question, we compare \sessionlearning with the previously discussed baselines. The results of this comparison are presented in Table~\ref{tab:main-results}.
The results demonstrate that \sessionlearning surpasses the baselines for 13 out of 18 agents, with 6 of these differences being statistically significant. Moreover, \sessionlearning significantly enhances the average performance across all agents compared to the baseline methods. Another notable observation is that \sessionlearning enhances the performance of \rlem. Although this improvement is not statistically significant at the 95\% confidence level, the p-value is quite small ($\text{p-value} = 0.061$), indicating a strong trend towards better performance. This observation suggests that applying \sessionlearning after \rlem is a promising method for the task at hand, indicating potential for further enhancing the system's performance.

\subsubsection*{\textbf{How does the batch size $b$ in \sessionlearning affects the performance?}}

One of the key hyperparameters in \sessionlearning is the batch size $b$. This parameter determines how many consecutive queries the search engine should serve using the same set of parameters, essentially defining the feedback collection interval before the model updates its parameters for the next batch. Exploring the impact of this parameter is insightful, as shown in Figure~\ref{fig:iteration-perfromance} (b) for average performance and Figure~\ref{fig:online-learning-batch-tasks} for per agent performance. While the optimal batch size might vary depending on the agent, the general trend suggests that increasing the batch size improves performance up to a certain point, beyond which performance starts to decline. Notably, setting a small value (e.g., 64) leads to a performance drop compared to the starting checkpoint (i.e., the checkpoint from the \rlem approach). We believe this occurs because using a small batch size $b$ leads to frequent updates in the model's parameters, making the model more susceptible to noisy feedback from the agent. These frequent updates might cause the model to overfit to specific feedback points rather than capturing generalizable patterns, thus negatively affecting performance.

\subsubsection*{\textbf{Does ``personalization'' with \ourframework result in different retrieval lists for different RAG agents?}}

We compare the retrieval lists for the same query across different agents when applying \ourframework to personalize the search engine for them. To evaluate the similarity between the retrieval lists, we employ two metrics: Kendall's tau coefficient and Jaccard's similarity. Kendall's tau coefficient captures how the ranking order of documents correlates between two lists, providing insights into how similarly or differently the documents are ranked for various agents. Jaccard's similarity, on the other hand, is a set-based metric that quantifies the overlap between two sets of retrieved documents, indicating the percentage of shared documents across retrieval lists for different agents.
These metrics allow us to analyze both the ranking and the content of the retrieved documents across personalized search engine settings.

The results of this experiment are shown in Figure~\ref{fig:ret-corr}, yielding several key insights. First, the findings suggest that, on average, about 20\% of the retrieved documents differ between RAG agents for the same query. This highlights the personalization effect, where each RAG agent receives a distinct set of documents despite querying with identical input. Furthermore, the low Kendall's tau correlation indicates significant differences in the ranking of the documents retrieved for different RAG agents, demonstrating that the search engine adapts document ranking based on agent-specific preferences and behaviors. Additionally, the Jaccard's similarity between agents employing FiD and those using in-prompt augmentation is notably lower than between agents both utilizing in-prompt augmentation. This demonstrates that the system has effectively learned to tailor document retrieval strategies according to the different retrieval-augmentation methods used by the RAG agents.

Another interesting observation is that the Kendall's tau correlation is higher between agents that both using T5 or FiD with T5 than between agents where one employs BART and the other utilizes T5 or FiD with T5. This suggests that the search engine, through model ID, has identified shared information needs between agents that utilize the same backbone language model (T5), resulting in more similar retrieval outputs. In summary, these results confirm that personalization significantly affects the retrieval lists provided to each agent, as the system learns to adapt document rankings and selections based on both the retrieval-augmentation method and the backbone LLM.

\section{Conclusion \& Future Work}

In this paper, we address the challenge of building a search engine tailored for multiple RAG agents, functioning similarly to how search engines serve human users, considering the paradigm shift where nowadays LLMs are the main users of the search engines. We propose \ourframework, an iterative framework with expectation maximization to iteratively gather feedback from RAG agents and adjust the search engine based on them in an offline and online phase. Our findings demonstrate that the proposed approach statistically significantly outperforms established baselines in terms of average agents performance. We also conducted extensive studies to analyze the impact of key factors such as the number of training iterations in \rlem, batch size in \sessionlearning, and the role of personalization in search results for each agent. Overall, the proposed method exhibits promising results, showcasing its effectiveness in designing search engines for multiple RAG agents.

There are several potential future directions for this work: (1) extending the current setup to optimize retrieval models rather than just reranking; (2) considering multiple utility functions per agent; (3) investigating novel regularization techniques to enhance generalization for agents who do not participate in training; (4) exploring interleaving and counterfactual learning approaches within the context of a search engine for machines; and (5) expanding beyond text generation to address a more general REML scenario.

\section*{Acknowledgments}
This work was supported in part by the Center for Intelligent Information Retrieval, in part by NSF grant number 2402873, and in part by the Office of Naval Research contract number N000142412612. Any opinions, findings and conclusions or recommendations expressed in this material are those of the authors and do not necessarily reflect those of the sponsors.

\balance
\bibliographystyle{ACM-Reference-Format}
\bibliography{XX-references}

\end{document}